\documentclass[letterpaper]{article} 
\usepackage{aaai25}  
\usepackage{times}  
\usepackage{helvet}  
\usepackage{courier}  
\usepackage[hyphens]{url}  
\usepackage{graphicx} 
\usepackage{natbib}  
\usepackage{caption} 
\usepackage{algorithm}
\usepackage{algorithmic}
\usepackage{amsmath}
\usepackage{amssymb}
\usepackage{multirow}
\usepackage{graphicx}
\usepackage{makecell}
\usepackage{booktabs}  
\usepackage{adjustbox}
\usepackage{array}
\usepackage{pifont}       
\usepackage{bbding}       
\usepackage{fontawesome}  
\usepackage{adjustbox}
\usepackage{amssymb}
\usepackage{natbib}

\floatstyle{ruled}
\newfloat{listing}{tb}{lst}{}
\floatname{listing}{Listing}
\usepackage{float}
\usepackage{pifont}
\usepackage{multirow}
\usepackage{booktabs}
\usepackage{newtxtext}
\usepackage{bm}
\usepackage{float}
\usepackage{subfigure}
\usepackage{multicol}
\usepackage{pdfpages}
\usepackage{indentfirst}
\usepackage{diagbox}
\usepackage{amsmath}
\usepackage{amsthm}
\usepackage{newfloat}
\usepackage{listings}

\usepackage{newfloat}
\usepackage{listings}
\DeclareCaptionStyle{ruled}{labelfont=normalfont,labelsep=colon,strut=off} 
\floatstyle{ruled}
\newfloat{listing}{tb}{lst}{}
\floatname{listing}{Listing}
\title{Spike2Former: Efficient Spiking Transformer for High-performance Image Segmentation}
\author{
    Zhenxin Lei\textsuperscript{\rm 1, \rm 2}\equalcontrib,
    Man Yao\textsuperscript{\rm 2}\thanks{Corresponding author.}, 
    Jiakui Hu\textsuperscript{\rm 3, \rm 2}\equalcontrib,
    Xinhao Luo\textsuperscript{\rm 2},
    Yanye Lu\textsuperscript{\rm 3, \rm 4},
    Bo Xu\textsuperscript{\rm 2},
    Guoqi Li\textsuperscript{\rm 2}\footnotemark[2]
}
\affiliations{
    \textsuperscript{\rm 1}University of Chinese Academy of Sciences\\
    \textsuperscript{\rm 2}Institute of Automation, Key Laboratory of Brain Cognition and Brain-inspired Intelligence Technology, \\ Chinese Academy of Sciences\\
    \textsuperscript{\rm 3}Institute of Medical Technology, Peking University Health Science Center, Peking University\\
    \textsuperscript{\rm 4}National Biomedical Imaging Center, Peking University
    
}

\begin{document}

\maketitle
\begin{abstract}
Spiking Neural Networks (SNNs) have a low-power advantage but perform poorly in image segmentation tasks. The reason is that directly converting neural networks with complex architectural designs for segmentation tasks into spiking versions leads to performance degradation and non-convergence. To address this challenge, we first identify the modules in the architecture design that lead to the severe reduction in spike firing, make targeted improvements, and propose Spike2Former architecture. Second, we propose normalized integer spiking neurons to solve the training stability problem of SNNs with complex architectures. We set a new state-of-the-art for SNNs in various semantic segmentation datasets, with a significant improvement of $\mathbf{+12.7\%}$ mIoU and $\mathbf{5.0\times }$ efficiency on ADE20K, $\mathbf{+14.3\%}$ mIoU and $\mathbf{5.2\times }$ efficiency on VOC2012, and $\mathbf{+9.1\%}$ mIoU and $\mathbf{6.6 \times }$ efficiency on CityScapes. Our code is available at https://github.com/BICLab/Spike2Former
\end{abstract}

%

\section{Introduction}

Spiking Neural Networks (SNNs) emulate the spatiotemporal dynamics and spike-based communication of biological neurons. The former ensures the network's representation \cite{1-3}, while the spike-driven paradigm introduced by the latter allows SNNs to perform sparse computing when deployed on neuromorphic chips \cite{1-5,1-4}, thereby benefiting from low power consumption \cite{1-1, 1-2}. A notable example is the sensing-computing neuromorphic chip Speck, which operates at a power level as low as 0.7mW in typical visual scenes \cite{Speck}.

The complex neuronal dynamics and binary activations make it challenging to train large-scale SNNs. It took a long time for the SNN field to effectively address this issue through surrogate gradient training \cite{Directtrainsnn1,neftci2019surrogate} and residual learning design \cite{SEW_resnet,MS-resnet}. Currently, SNNs have achieved commendable performance on simple image classification tasks \cite{Spikeformer,Spikeformerv2}. Unfortunately, when it comes to complex visual tasks that require the use of sophisticated neural network architectures, SNNs fall short.

For instance, in image segmentation, an additional segmentation head is required alongside the backbone used for image classification, resulting in a network structure that is significantly more complex than that for classification. Simply converting complex Artificial Neural Networks (ANNs) into spiking versions or directly applying residual designs from classification tasks often leads to a notable drop in performance. Consequently, the few existing SNN models \cite{kimUNet,SCGNet,Spikeformerv2,su2024multi,SpikeUNET} that tackle image segmentation tasks tend to perform poorly.

This work explores the application of SNNs with more complex architectures to image segmentation tasks. Specifically, Mask2Former \cite{Mask2Former} is a classic Transformer-based per-mask classification architecture consisting of three parts: a backbone network; a Feature Pyramid Network (FPN) pixel decoder with a multi-scale deformable transformer encoder block; and a transformer decoder block. Directly converting the Mask2Former architecture into a spiking version results in obvious performance degradation and non-convergence. To address this, we investigated which modules in the spiking Mask2Former contribute to the significant loss of information, where the spiking neurons in these modules nearly cease to fire.

The first module with severe spike degradation is the deformable attention transformer encoder block in the FPN decoder. In the vanilla Mask2Former, the query operation in the deformable attention block is inherently sparse; if the queried information consists of sparse spikes, this could result in significant information loss. To address this, we incorporate convolution blocks in encoder blocks and redesign the deformable attention blocks to preserve more effective information and energy efficiency. The second key module that leads to information loss is the final mask embedding layer. This layer is crucial as it outputs the final segmentation results; however, being at the deepest part of the network, the semantic information that reaches this point is already greatly diminished. To end this, we build an auxiliary information branch to enhance the representation of mask embedding.

Another challenge is that, beyond architectural design, spiking neurons inherently introduce information loss by converting continuous values into binary spikes. This issue has long plagued the SNN field, leading to the development of various methods aimed at refining the membrane potential distribution, such as attention mechanism \cite{attentionSNN} and information maximization loss \cite{informationmax}. Recently, \citeauthor{Xinhao} proposed an integer training and spike inference method in spiking CNNs to address this challenge. However, this approach does not extend to more complex architectures like Mask2Former, which require numerical stability and precise representation in the interaction of cross-modal features \cite{Maskformer,DETR,maskdino}. To address this, we propose a novel normalization method that normalizes integers during training to facilitate training, while ensuring that the spike-driven nature during inference remains unaffected.

The proposed methods are validated on the popular datasets ADE20k, CityScapes, Pascal VOC2012 for semantic segmentation. Spike2Former far exceeds the best models in SNNs in both performance and energy consumption and is comparable to the ANNs. The main contributions are:

1) \textbf{Spike2Former}. We analysis the challenge of applying SNNs to complex architecture and propose a transformer-based image segmentation method, Spike2Former, that incorporates the Spike-driven Deformable Transformer Encoder (SDTE) and Spike-Driven Mask Embedding (SDME) module to resolve the information deficiency and improve model performance. 

2) \textbf{NI-LIF Spiking Neuron}. We design a novel spiking neuron, NI-LIF, to increase the training stability and reduce the information deficiency in complex segmentation method. The NI-LIF normalizes the integer activation during training and can be equivalent to spike-driven in inference, capitalizing on the low power nature of SNNs.

3) \textbf{Performance}: The proposed Spike2Former obtains a remarkable performance improvement in popular segmentation tasks with low power consumption, demonstrating the potential of SNNs in complex scenarios. Spike2Former achieves a new state-of-the-art ADE20k ($\mathbf{+12.7\% }$ mIoU and $\mathbf{5.0\times }$), Pascal VOC2012 ($\mathbf{+14.3\% }$ mIoU and $\mathbf{5.2\times }$), CityScapes ($\mathbf{+9.1\% }$ mIoU and $\mathbf{6.6\times }$).
\vspace{-0.2cm}

\begin{figure*}[t]
\centering
\includegraphics[width=1.0\textwidth]{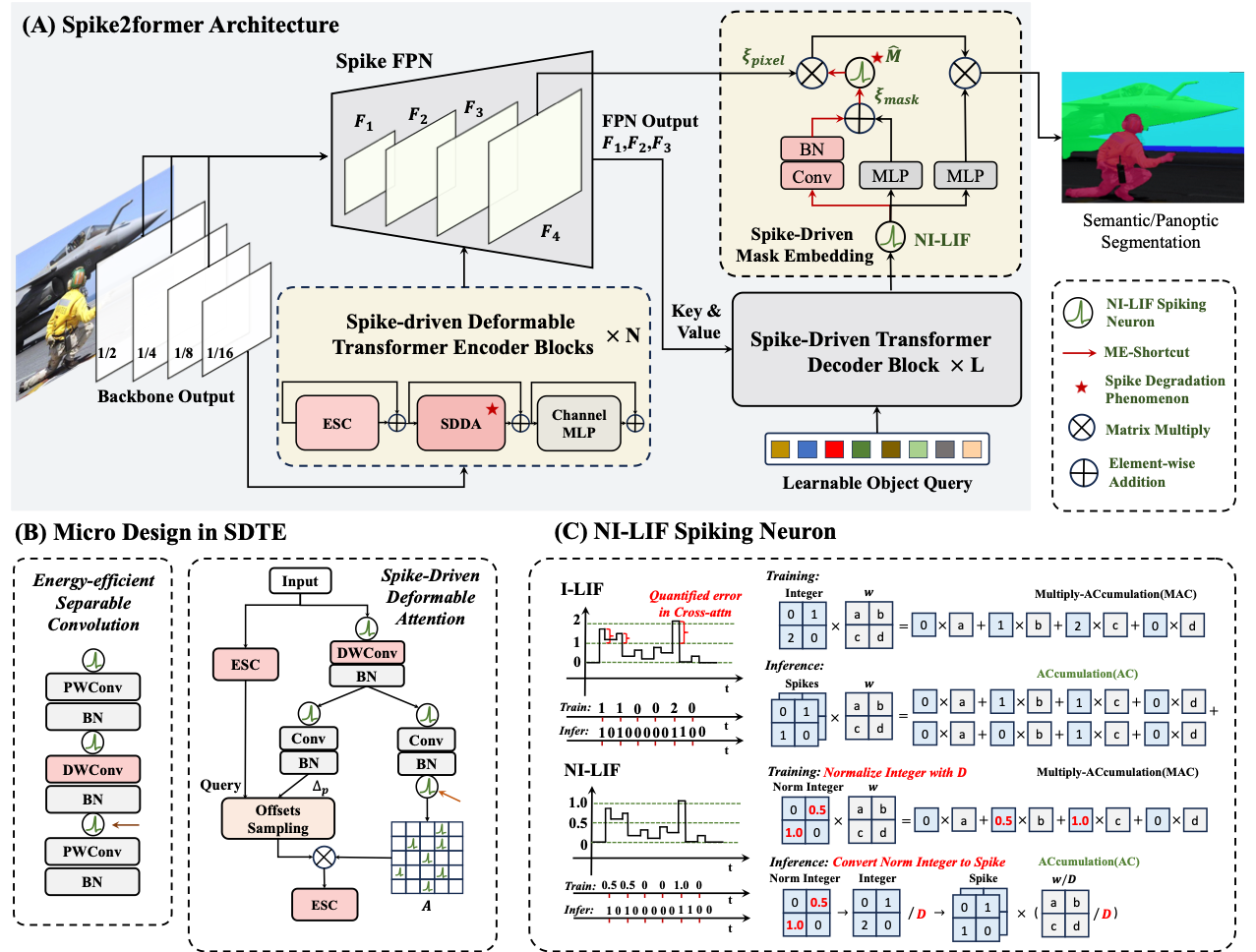} 
\caption{(A) The architecture of Spike2Former, we propose Spike-Driven Mask Embedding(SDME) and Spike-driven Deformable Transformer Encoder to introduce the transformer-based method to SNNs. (B) Micro Design in Spike-Driven Deformable Transformer Encoder(SDTE) including Spike-driven Deformable attention(SDDA) and Spike Separable Convolution(ESC). (C) Comparison between NI-LIF and I-LIF. Integer activation results in information loss, especially in cross-model interaction. NI-LIF normalizes the integer during training to preserve information and converts them to spikes during inference with only sparse addition.}
\label{fig1} 
\vspace{-0.2cm}
\end{figure*}

\section{Related Works}
\subsection{SNN Training}
Training methods in Spiking Neural Networks (SNNs) mainly fall into two categories: ANN-to-SNN conversion and direct training. While ANN-to-SNN conversion inherits from traditional neural networks, it suffers from longer time steps and limited real-time processing capabilities \cite{spikecalibration, ANN2SNN, ANN2SNN2021}. Direct training with surrogate gradients \cite{surrogategradient} offers better flexibility but typically yields lower performance. Inspired by recent Integer-based Leaky Integrate-and-Fire (I-LIF) neurons, which bridge the gap between training and inference through virtual timesteps, we adopt the direct training approach for its architectural flexibility \cite{MS-resnet, SEW_resnet, Directtrainsnn1}.
Recently, \citeauthor{Xinhao} introduced a novel spiking neuron that trains the model with Integer activation as a virtual timesteps and converts the Integer into binary spikes by expanding virtual timesteps during inference. We take this inspiration and further improve it for more complex application scenes.
\vspace{-0.2cm}

\subsection{SNN Backbone Design}
SNN backbone designs can be broadly categorized into CNN-based and Transformer-based approaches. CNN-based SNNs primarily focus on improving the Spike-ResNet architecture \cite{vanillaSpikeresnet} through variations such as SEW-ResNet \cite{SEW_resnet} and MS-ResNet \cite{MS-resnet}. These variations utilize diverse residual connections to alleviate performance degradation and support deeper architectures. Recently, Transformer-based methods have gained prominence in SNN backbone design. Approaches such as \cite{spiketransformer111, spiketransformer222ZZK, spiketransformer333, Spikeformer, Spikeformerv2, hu2024highperformance, spikeformerv3} incorporate spiking neurons into self-attention mechanisms to enhance model performance. These methods are relatively straightforward and lack complex interactions, making them easier to train within the SNN framework. Additionally, \citeauthor{Xinhao} recently introduced a spike-driven SpikeYOLO model, adapted from YOLOv8 for object detection, which achieves competitive performance. However, to mitigate information loss during feature interactions, this method simplifies the model design, failing to fully address the challenges of applying SNNs to complex architectures.

\subsection{Image Segmentation}

In ANNs, image segmentation can be categorized into per-pixel and per-mask classification. Per-pixel classification, often using CNNs, employs Feature Pyramid Networks (FPN) to generate segmentation masks. MaskFormer \cite{Maskformer} redefined this as per-mask classification, generating binary masks and assigning them semantic classes. Mask2Former \cite{Mask2Former} and MaskDINO \cite{maskdino} further improve the MaskFormer with more precise refinement of learnable query. In SNNs, image segmentation remains challenging \cite{spikecalibration, spikeseg, SpikeUNET}. In 2022, \cite{kimUNet} introduced spike-driven decoders (spike-FCN and spike-Deeplab) by converting ANN architectures to SNNs, while Spiking-CGNet \cite{SCGNet} in 2023 developed a segmentation-specific backbone. More recently, Meta-SpikeFormer \cite{Spikeformerv2} directly trained on FPN achieved competitive results on ADE20k \cite{ADE20k}, highlighting SNN potential. However, the performance and energy efficiency of these method are poor and limit their application to various scenarios.
\vspace{-0.2cm}

\section{Method}
Spike2Former adopts the architecture of Mask2Former, which includes a pyramid backbone, a pixel decoder with a deformable transformer encoder, a transformer decoder, and a mask embedding module. In this section, we present the improvement in the deformable transformer encoder and mask embedding module, followed by a discussion of the newly proposed spiking neuron NI-LIF.

\subsection{Information Deficiency in Query}
Query features are essential in transformer-based methods \cite{dropqueries, mixquerytransformer, YouOnlyLearnOneQuery}. They are primarily used to interact with image features in the Multi-Scale Deformable Attention (MSDeformAttn) transformer encoder and transformer decoder within Mask2Former. However, SNNs face challenges in refining queries due to the significant loss of rich semantic representations when converting query into binary spikes. This raise the question: \textit{How can queries preserve information effectively in SNNs?} To address this, we propose modifications in the architecture design, including deformable attention blocks and the mask embedding module, which are particularly prone to bias in SNNs. The details of these modifications will be discussed in the following of this section.

\subsubsection{Spike-driven Deformable Transformer Encoder}
Spike-driven Deformable Transformer Encoder(SDTE) consists of a stack of successive blocks, including an Energy-efficient Separable Convolution(ESC) module to enhance local connections, a Spike-Driven Deformable Attention(SDDA) module that conducts deformable attention within queries, and a Channel-MLP layer to learn non-linear representations.

\noindent \textbf{Energy-efficient Separable Convolution Blocks} \citeauthor{Spikeformerv2} utilize a separable convolution to enhance the inductive bias. However, this design significantly increases energy consumption due to the direct connection of depthwise and pointwise convolutions. Thus, we propose adding spiking neuron before the second pointwise convolution for energy efficiency, denoted as $ESC(\cdot)$, and formulated as follows:
\begin{align}
     \mathbf{U}_{pw1}  &= {\rm Conv}_{pw1}(SN(\mathbf{U}))     \\
     \mathbf{U}_{dw}   &= {\rm Conv}_{dw}(SN(\mathbf{U}_{pw1}))  \\
     \mathbf{U}_{pw2}  &= {\rm Conv}_{pw2}(SN(\mathbf{U}_{dw}))
\end{align}
where $\mathbf{U}\in\mathbb{R}^{\frac{H}{16}\times \frac{W}{16}\times dim}$ means the input membrane potential ($dim$ denotes the embedding dimension), ${\rm Conv}_{dw}(\cdot)$ denotes depthwise convolutions, and ${\rm Conv}_{pw}(\cdot)$denote pointwise convolutions.

\begin{figure}[!t]
\centering
\includegraphics[width=0.98\linewidth]{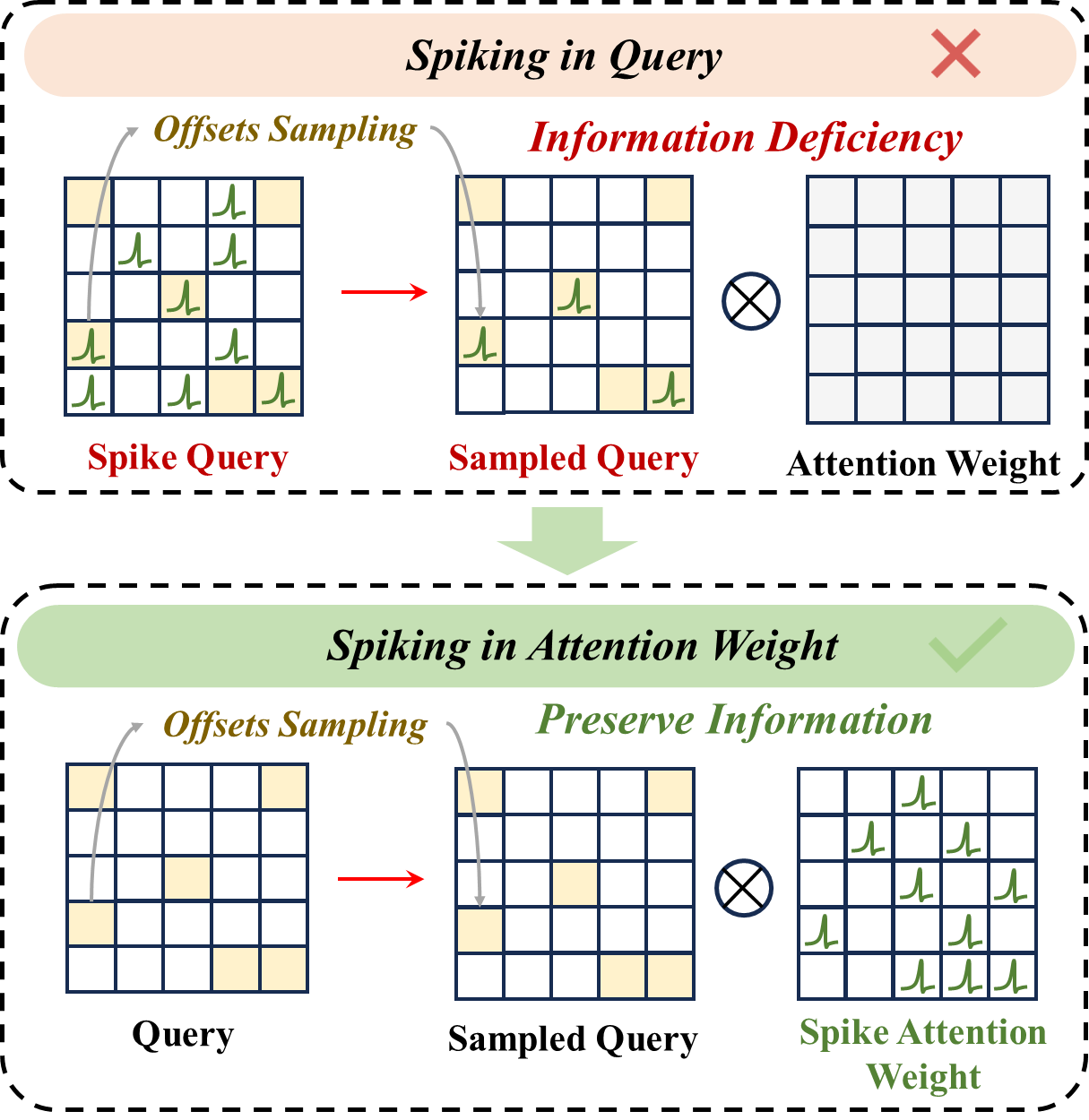} 
\caption{Illustration of offset-sampling operation in Spike-Driven Deformable Attention (SDDA). While directly spiking the query (Sampled Spike Query) leads to information loss during attention, spiking the attention weights (Spike Attention Weight) effectively preserves this crucial query information.}
\label{fig2} 
\vspace{-0.2cm}
\end{figure}

\noindent \textbf{Spike-Driven Deformable Attention} The MSDeformAttn transformer encoder in Mask2Former refines features by attending to the global context, dynamically computing weights from the inputs, and utilizing deformability to adapt the receptive field size \cite{PEMMaskFormer}. Although effective in ANNs, the sparse sampling strategy in deformable attention leads to information loss and unstable gradients (Fig. \ref{fig2}). Thus, we propose converting the attention weights into spikes instead of spiking the feature queries to better preserve the semantic information contained within the queries. Furthermore, using multi-scale features as queries significantly increases energy consumption, especially with high-resolution inputs. We recommend using single-scale deep image features with convolution blocks to enhance local connectivity, thereby improving energy efficiency while maintaining performance.

Specifically, for the calculation of attention weight and sampling offsets, we propose adding depthwise convolution (DWConv) to enhance the understanding of scene context. As shown in Fig. \ref{fig1}(B), given an input feature map $x_g$, the Spike-Driven Deformable Attention can be formulated as:
\begin{align}
{\rm DeformableSDSA}(\mathbf{p_q}, \mathbf{x_g}) = \sum_{g=1}^{G} \sum_{k=1}^{K} \mathbf{W}_g \mathbf{A}_{gk} \mathbf{W}^{'}_{g} \\ \nonumber
\cdot \mathbf{x}_g(p_0 + p_k + \Delta p_{gk})
\end{align}
where $G$ represents the total number of aggregation groups. For the $g$-th group, $\mathbf{W}g$ and $\mathbf{W}^{'}{g}$ represent the location-irrelevant projection weights. $\mathbf{A}{gk}$ denotes the attention weight corresponding to the $k$-th sampling point within the $g$-th group. $\Delta p{gk}$ represents the offset for the $k$-th sampling location $p_k$ within the $g$-th group. Subsequently, $\mathbf{W}g$, $\mathbf{A}{gk}$, and $\Delta p_{gk}$ are calculated as follows:
\begin{align}
\mathbf{W}g &= {\rm ESC}(x_g), \\
x_g^{'} &= {\rm BN}({\rm DWConv}(SN(x_g))), \\
\mathbf{A}{gk} &= SN({\rm BN}({\rm Conv}(SN(x_g^{'})))), \\
\Delta p_{gk} &= {\rm BN}({\rm Conv}(SN(x_g^{'}))).
\end{align}
Where the $SN$ represents the spiking neuron. We add spiking neuron for $\mathbf{A}{gk}$ to convert the attention weight into spike and maintain the effective information in feature query.
Finally, we apply an ESC block to the input that has been sampled to the embedding dimension. The output from the SDTE is then fed into the SpikeFPN \cite{Spikeformerv2} to generate per-pixel embedding.

\begin{table*}[!t]
\centering

\begin{adjustbox}{max width=1.05\linewidth}
\begin{tabular}{cllcccc}
\toprule[1pt]

\multicolumn{2}{c}{Method}               & \multicolumn{1}{c}{Model}                                                               & \begin{tabular}[c]{@{}c@{}}mIoU\\ (\%)\end{tabular}                 & \begin{tabular}[c]{@{}c@{}}Params\\ (M)\end{tabular} & $T \times D$ & \begin{tabular}[c]{@{}c@{}}Power\\ (mJ)\end{tabular} \\ 
\toprule[0.5pt]
\multicolumn{7}{c}{\textbf{Pascal VOC2012}}                                                         \\ \bottomrule[0.5pt]
\multicolumn{2}{c}{\multirow{3}{*}{ANN}} & \multicolumn{1}{l}{FCN\cite{FCN}}                                                                 & 62.2                 & 49                                                   & -            & 909.6                                                \\
\multicolumn{2}{c}{}                     & \multicolumn{1}{l}{DeepLabv3\cite{deeplab}}                                                           & 66.7                 & 68                                                   & -            & 1240.6                                               \\
\multicolumn{2}{c}{}                     & \multicolumn{1}{l}{DeepLabv3+\cite{Deeplabv3}}                                                          & 77.2                 & 41                                                   & -            & 326.6                                                \\ 
\midrule
\multicolumn{2}{c}{ANN2SNN}              & \multicolumn{1}{l}{Spike Calibration\cite{spikecalibration}}                                                   & 59.6                   & -                                                    & 256$\times$1           & -                                                    \\ 
\midrule
\multicolumn{2}{c}{\multirow{7}{*}{SNN}} & \multicolumn{1}{l}{SpikeFCN\cite{kimUNet}}                                                            & 9.9                  & 50                                                   & 20$\times$1            & 383.5                                                \\
\multicolumn{2}{c}{}                     & \multicolumn{1}{l}{SpikeDeepLab\cite{kimUNet}}                                                        & 22.3                 & 68                                                   & 20$\times$1           & 523.2                                                \\
\multicolumn{2}{c}{}                     & \multicolumn{1}{l}{\multirow{2}{*}{SpikeFPN\cite{Spikeformerv2}}}                                           & 58.1                 & 17                                                   & 1$\times$1     & 81.4                                                 \\
\multicolumn{2}{c}{}                     & \multicolumn{1}{l}{}                                                                    & 61.1                 & 59                                                   & 4$\times$1     & 179.4                                                \\ \cline{3-7}
\multicolumn{2}{c}{}                     & \multicolumn{1}{l}{\multirow{4}{*}{\textbf{Spike2Former  (Ours)}}} & 61.8                     & 34                                                   & 1$\times$2    &  50.6                                                    \\

\multicolumn{2}{c}{}                     & \multicolumn{1}{l}{}                                                                    &  62.1         & 34                                                   & 2$\times$2    &  98.3                                                    \\
\multicolumn{2}{c}{}                     & \multicolumn{1}{l}{}                                                                    & 75.1          & 34                                                   & 1$\times$4    & 63.0                                                      \\
\multicolumn{2}{c}{}                     & \multicolumn{1}{l}{}                                                                    & \textbf{75.4} & 34                                                   & 4$\times$4     &  221.9                                     \\ 
\toprule[0.5pt]
\multicolumn{7}{c}{\textbf{ADE20k}} \\ 
\bottomrule[0.5pt]
\multicolumn{2}{c}{\multirow{5}{*}{ANN}} & \multicolumn{1}{l}{DeepLabv3+\cite{Deeplabv3}}                                                          & 42.7                 & 41                                                   & -            & 818.8                                                \\
\multicolumn{2}{c}{}                     & \multicolumn{1}{l}{\multirow{2}{*}{MaskFormer\cite{Maskformer}}}                                         & 44.5                 & 41                                                   & -            & 243.8                                                \\
\multicolumn{2}{c}{}                     & \multicolumn{1}{l}{}                                                                    & 46.7                 & 42                                                   & -            & 253                                                  \\
\multicolumn{2}{c}{}                     & \multicolumn{1}{l}{\multirow{2}{*}{Mask2Former\cite{Mask2Former}}}                                        & 47.2                 & 44                                                   & -            & 326.6                                                \\
\multicolumn{2}{c}{}                     & \multicolumn{1}{l}{}                                                                    & 47.7                 & 47                                                   & -            & 340.6                                                \\ 
\midrule
\multicolumn{2}{c}{\multirow{2}{*}{SNN}} & \multicolumn{1}{l}{SpikeFPN\cite{Spikeformerv2}}                                                            & 33.6                 & 17                                                   & 4$\times$1     & 88.1                                                 \\ 
\multicolumn{2}{c}{}                     & \multicolumn{1}{l}{\textbf{Spike2Former  (Ours)}}                  & \textbf{46.3} & 34                                                   & 1$\times$4     & 68.0                                        \\ 
\midrule[0.5pt]
\multicolumn{7}{c}{\textbf{CityScapes}}                                                          \\ \bottomrule[0.5pt]
\multicolumn{2}{c}{\multirow{5}{*}{ANN}} & DeepLabv3+\cite{Deeplabv3}                         & 79.1                                                & 65                                                   & -            & 1241.3                                               \\
\multicolumn{2}{c}{}                     & \multirow{2}{*}{MaskFormer\cite{Maskformer}}       & 78.5                                                & 41                                                   & -            & 463.5                                                \\
\multicolumn{2}{c}{}                     &                                                    & -                                                   & -                                                    & -            & -                                                    \\
\multicolumn{2}{c}{}                     & \multirow{2}{*}{Mask2Former\cite{Mask2Former}}     & 79.6                                                & 44                                                   & -            & 601.6                                                \\
\multicolumn{2}{c}{}                     &                                                    & 80.6                                                & 47                                                   & -            & 623.4                                                \\ \midrule
\multicolumn{2}{c}{\multirow{2}{*}{SNN}} & \multicolumn{1}{l}{SCGNet-L\cite{SCGNet}}                                                            & 66.5                 & 1.9                                                  & 4$\times$1            & 24.6                                                 \\ 
\multicolumn{2}{c}{}                     & \multicolumn{1}{l}{\textbf{Spike2Former (Ours)}}                  & \textbf{75.2}  & 34                                                   & 1$\times$4     & 93.8                                                 \\ 
\bottomrule[1pt]
\end{tabular}%
\end{adjustbox}
\caption{The semantic segmentation results on ADE20k, CityScapes, and Pascal VOC2012 are presented. We obtain $+14.3\%$,$+12.7\%$, and $+9.1\%$mIoU on Pascal VOC, ADE20k, and CityScapes separately compared with the state-of-the-art SNN method. The results of MaskFormer and Mask2Former are presented with R50 (Above) and Swin-t (Below) as backbone, respectively. Power is the estimation of energy consumption as formulated in \cite{Spikeformer}.}
\label{tab:segmentation}
\vspace{-0.4cm}
\end{table*}

\subsubsection{Spike-Driven Transformer Decoder}
The Spike-Driven Transformer Decoder (SDTD) contains a Spike-Driven Cross-Attention (SDCA) layer, a Spike-Driven Self-Attention (SDSA) layer, and a Channel-MLP layer. The formulation of SDTD can be written as:
\begin{align}
\mathbf{Q}^{'} &= \mathbf{Q} + {\rm SDCA}(\mathbf{Q}, \mathbf{F_i}), i=1,2,3 \\
\mathbf{Q}^{''} &= \mathbf{Q}^{'} + {\rm SDSA}(\mathbf{Q}^{'}), \\
\mathbf{Q}^{'''} &= \mathbf{Q}^{''} + {\rm ChannelMLP}(\mathbf{Q}^{''}). 
\end{align}
where $\mathbf{Q}\in\mathbb{R}^{N\times C}$ are the $N$ learnable query with with learnable positional embedding and $\mathbf{F_i} \in \mathbb{R}^{H_i \times W_i\times B_i}$ indicate the multi-scale feature maps obtained from the pixel decoder, with $i\in \{1,2,3\}$.

Additionally, we replace the re-parameterization convolution \cite{Spikeformerv2} with the combination of Linear and BatchNorm for energy efficiency. The formulation of SDSA can be written as:
\begin{align}
\mathbf{Q_s}, \mathbf{K_s}, \mathbf{V_s} &= SN({\rm BN}({\rm Conv}(SN(\mathbf{Q_{l-1}})))),  \\ 
\mathbf{Q_l} &= {\rm BN}({\rm Conv}(SN(\mathbf{Q_s}\mathbf{K_s}^T\mathbf{V_s} \ast Scale)) 
\end{align}
where the $scale$ of SDSA can be re-parameterized into the spiking neuron's threshold. Note that the SDCA obtains the $K_s$ and $V_s$ from multi-scale feature maps $\mathbf{F_i}$.

\subsubsection{Spike-Driven Mask Embedding}
The mask embedding module in Mask2Former \cite{Mask2Former} uses a Multi-Layer Perceptron (MLP) to convert the per-segment embedding $\mathbf{Q}$ to $N$ mask embedding $\mathbf{\zeta}_{mask} \in 
\mathbb{R}^{N \times C}$, where C denotes the object class and N denotes the number of query and obtain the binary mask prediction $\mathbf{\hat{M}} \in [0, 1]$ through dot product between per-segment embedding $M$ and per-pixel embedding $\mathbf{\zeta}_{pixel}$. However, the rich diversity semantic information corresponding to the segment object within $N$ mask embedding $\mathbf{\zeta}_{mask}$ will suffer a significant reduction when converting to binary spikes and further affect the binary mask predictions. Furthermore, as shown in Fig. \ref{fig3} we also found the spike degradation phenomenon in mask embedding that the spiking firing rate before the embedding operation suffers a significant loss and infect the performance. Thus, we proposed a Membrane Embedding Shortcut (ME-Shortcut) to connect the membrane potential from the output of the transformer decoder to mask embedding, which significantly ensures the training stability and enhances the effective representation of $N$ mask embedding $\mathbf{\zeta}_{mask}$.

As shown in Fig. \ref{fig1}(A), we build a ME-Shortcut with a channel convolution \cite{mobilenetv3} parallel with the MLP layer to enhance the precise representation of $\mathbf{\zeta}_{mask}$ and reduce the spike degradation phenomenon. In the following, we convert the $\mathbf{\zeta}_{mask}$ to spikes before multiplying to pixel embedding $\mathbf{\zeta}_{pixel}$ for spike-driven. Finally, we obtain the binary mask predictions by:
\begin{align}
\mathbf{\zeta}_{mask} &= SN({\rm MLP}(SN(\mathbf{Q})) + w_s \ast {\rm BN}({\rm Conv}(SN(\mathbf{Q})))), \\
\mathbf{\hat{M}} &= \mathbf{\zeta}_{mask} \cdot \mathbf{\zeta}_{pixel}.
\end{align}
where the $SN(\cdot)$ is the spiking neuron layer, $w_s$ is a learnable weight initialized with 1 to select key feature.

\subsection{NI-LIF Spiking Neuron}
The spiking neuron layer integrates spatio-temporal information into the membrane potential and then converts it into binary spikes for spike-driven computing in the following layer. Different from the image classification task, dense prediction necessitates a higher demand for numerical stability. Recent work \cite{Xinhao} shows a performance increase in Object Detection with Integer Leaky Integrate-and-Fire (I-LIF). However, when we attempt to extend the I-LIF to a more complex architecture like Mask2Former, we found that gradient instability occurs, especially in the spike-driven transformer decoder layer. We argue that the potential reason is that the interaction between object queries and multi-scale feature maps requires numerical stability. However, the large integer value in multi-scale feature maps obtained from I-LIF failed to quantify the precise value and brought quantization error and information loss. Therefore, we intend to tackle this issue by normalizing the $S[t]$ with the virtual timesteps $D$ to maintain the gradient stable and preserve the information.

We improved I-LIF through normalizing integer-value with virtual timesteps to enhance the numerical and gradient stability, as depicted in Fig. \ref{fig1}(C), whose dynamics are:
\begin{align}
& U[t] = H[t-1] + X[t], \\
& S[t] = Clip(round(U[t]), 0, D) / D, \\
& H[t] = \beta(U[t] - S[t]\times D). 
\end{align}
where $X[t]$ is the spatial input current at timestep $t$, $U[t]$ denotes the membrane potential that integrates the temporal information $H[t-1]$ and spatial $X[t]$. $round(\cdot)$ is a round operation, $Clip(x, min, max)$ implies clipping the input $x$ to $[min, max]$, and $D$ is a hyper-parameter to emit the maximum integer value.

\noindent \textbf{Inference}
As shown in Fig. \ref{fig1}(C), I-LIF \cite{Xinhao} converts the integer spikes into binary spikes during inference stages, and the input to the spiking neuron at $l+1$ layer can be described as:
\begin{align}
X^{l+1}[t] = W^l S^{l} [t]
\end{align}
We also extend the T time step to $T\times D$, and convert the normalized value $S^{l}[t]$ to a spike sequence $\{S^{l}[t, d]\}^{D}_{d=1}$, which can be written as:
\begin{align}
\sum_{d=1}^{D} S^l[t, d] = S^l[t] \times D
\end{align}
Therefore, the input of the neuron at $l+1$ layer can be calculated by:
\begin{align}
    X^l[t] = \sum_{d=1}^{D} (W^l (S^l[t, d] \times D)), \\ \nonumber
    X^l[t] = \sum_{d=1}^{D} (W^{l}_D (S^l[t, d])).
\end{align}
where the $W^{l}_D=W^l \times D$. The spike sequence $S^l[t, d]$ only contains $0/1$ after being multiplied by the quantization step, and the MAC (Multiply-Accumulation operations) can be converted to sparse AC (Accumulation operations), which enables sparse addition during inference.

\section{Experiment}
\textbf{Dataset.} We conduct semantic segmentation on ADE20k \cite{ADE20k}, CityScapes \cite{cityscapes}, and Pascal VOC2012 \cite{pascal-voc-2012} datasets. The details of the training strategy are shown in Tab. \ref{tab:Dataset}.

\textbf{Training setting.} We present all our main semantic segmentation results in mean Intersection over Union (mIoU) under single scale inference setting. We use Meta-Spikeformer (Params:15M) \cite{Spikeformerv2} as our backbone, which is pre-trained on ImageNet-1k for 200 epochs. More training details can be found in the Appendix.
\begin{table}[htbp]
\centering
\begin{adjustbox}{max width=\linewidth} 
\begin{tabular}{lccc}
\toprule[1pt]
Setting & ADE20k & CityScapes & \begin{tabular}[c]{@{}l@{}}Pasval\\ VOC2012\end{tabular} \\ \midrule[0.5pt]
Input Size & 512$\times$512 & 512$\times$1024 & 512$\times$512 \\
Learning Rate & 2e-4 & 2e-3 & 2e-3 \\
Optimizer & AdamW  & AdamW & AdamW \\
Training Steps & 160k & 90k & 80k \\ \bottomrule[1pt]
\end{tabular}%
\end{adjustbox}
\caption{Hyper-parameters setting in Spike2Former.}
\label{tab:Dataset}
\vspace{-0.1cm}
\end{table}

\begin{table*}[!t]
\centering
\begin{adjustbox}{max width=1.2\linewidth} 
\tabcolsep=0.05cm
\begin{tabular}{cccc}
\toprule[1pt]
Ablation & Method & Power (mJ)& mIoU (\%)\\ \midrule[0.5pt]
 & \textbf{Spike2Former(Baseline)} & \textbf{68.0} & \textbf{46.3} \\ \midrule[0.5pt]
ANN Conversion & Mask2Former(Spiking Version) & - & 5.2 \\ \midrule[0.5pt]
\multirow{3}{*}{\begin{tabular}[c]{@{}c@{}}Spike-Driven \\ Mask Embedding\end{tabular}} & Vanilla Spike Mask Embedding & 67.9 & 41.8(-4.5) \\
 & w/o. Conv in ME-Shortcut & 67.9 & 45.7(-0.6) \\
 & ME-Shortcut $\rightarrow$ Membrane Shortcut & 67.8 & 45.0(-1.3) \\ \midrule[0.5pt]
\multirow{6}{*}{\begin{tabular}[c]{@{}c@{}}Spike-driven Deformable \quad \\ Transformer Encoder\end{tabular}} & w/o. SD Deformable Transformer Encoder & 65.2 & 43.8(-2.5) \\
 & w/o. ESC in SD Deformable attention & 64.5 & 45.6(-0.7) \\
 & w/o. DWConv in SD Deformable attention & 67.8 & 45.7(-0.6) \\ \cline{2-4}
 & Spiking in Query & 67.8 & 43.1(-3.2) \\
 & Multi-Scale feature as query & 136.5 & 46.7(+0.4) \\
 & Vanilla Transformer Encoder & 66.9 & 44.9(-1.4) \\ \midrule[0.5pt]
\multirow{4}{*}{\begin{tabular}[c]{@{}c@{}}Normalize Method \\ in NI-LIF\end{tabular}} & I-LIF & 68.1 & 37.9(-8.4) \\
 & Norm Only in Cross-Attn & 68.1 & 43.2(-3.1) \\ \cline{2-4}
 & Norm with Const 8 & 67.4 & 44.7(-1.3) \\
 & Norm with Sigmoid\&$L2$ & - & $*$ \\ \bottomrule[1pt]
\end{tabular}%
\end{adjustbox}
\caption{We conduct ablation on the proposed Spike-Driven Mask Embedding (SDME), Spike-driven Deformable Transformer Encoder (SDTE) and Normalized Integer-LIF (NI-LIF) in ADE20k dataset. We set $T\times D=1\times 4$ and modify just one modification of the baseline to test how the power and performance vary. * Does not converge. The ablation on ANN indicates the direct conversion of Mask2Former with SNNs.}
\label{tab:ablation}
\vspace{-0.4cm}
\end{table*}

\subsection{Experiment Results}
In Tab. \ref{tab:segmentation}, we comprehensively compare Spike2Former with other ANN and SNN methods in mIoU, parameters, and power. The proposed Spike2Former significantly improves the performance upper bound of SNNs on three public datasets. We obtain \textbf{46.3\% mIoU}, \textbf{75.2\% mIoU}, and \textbf{75.4\% mIoU} in ADE20k, CityScapes, and Pascal VOC2012 which is $\mathbf{+14.3\%}$, $\mathbf{+9.1\%}$, and $\mathbf{+14.3\%}$ higher than the previous state-of-the-art SNN method \cite{Spikeformerv2, SCGNet}, respectively. Spike2Former also demonstrates significant advantages over the existing SNNs in terms of energy consumption: Spike2Former vs. SpikeFPN \cite{Spikeformerv2}: mIoU \textbf{44.6\%} vs. 33.6\%; Power: \textbf{68mJ} vs. 88.1mJ in ADE20k dataset and Spike2Former vs. SpikeFPN: mIoU \textbf{75.4\%} vs. 61.1\%; Power \textbf{63.0mJ} vs. 179.4mJ in Pascal VOC2012 dataset. Moreover, the performance gap between SNNs and ANNs is significantly narrowed. Spike2Former got $\mathbf{+1.2\%}$ mIoU compared with MaskFormer(R50) and is competitive with the current classical ANN architecture Mask2Former(R50) in ADE20k dataset, which is 46.3\% vs. 44.5\% vs. 47.2\% mIoU in mIoU while the energy consumption is much lower for $\mathbf{3.58\times}$ and $\mathbf{4.80\times}$ energy efficiency. 
\vspace{-0.1cm}

\subsection{Ablation study}
We conduct ablation studies on the various componects of Spike2Former to evaluate the contribution of each parts. 

\subsubsection{Spike-Driven Deformable Transformer Encoder}

Tab. \ref{tab:ablation} highlights the performance of our spike-driven Deformable Transformer Encoder (SDTE). As shown in Fig. \ref{fig2}, directly spiking query features reduces mIoU by 3.2\%, likely due to excess retention of attention weights, diminishing effective information of query. Our SDTE improves mIoU by 2.5\% over SpikeFPN \cite{Spikeformerv2} without Transformer encoder and by 1.1\% over a vanilla Transformer. While using multi-scale features as queries, as in Mask2Former, yields 46.7\% mIoU, it doubles energy consumption (68.0mJ vs. 136.5mJ), prompting us to prioritize single-scale queries for energy efficiency.

\begin{figure}[!t]
\centering 
\includegraphics[width=0.98\linewidth]{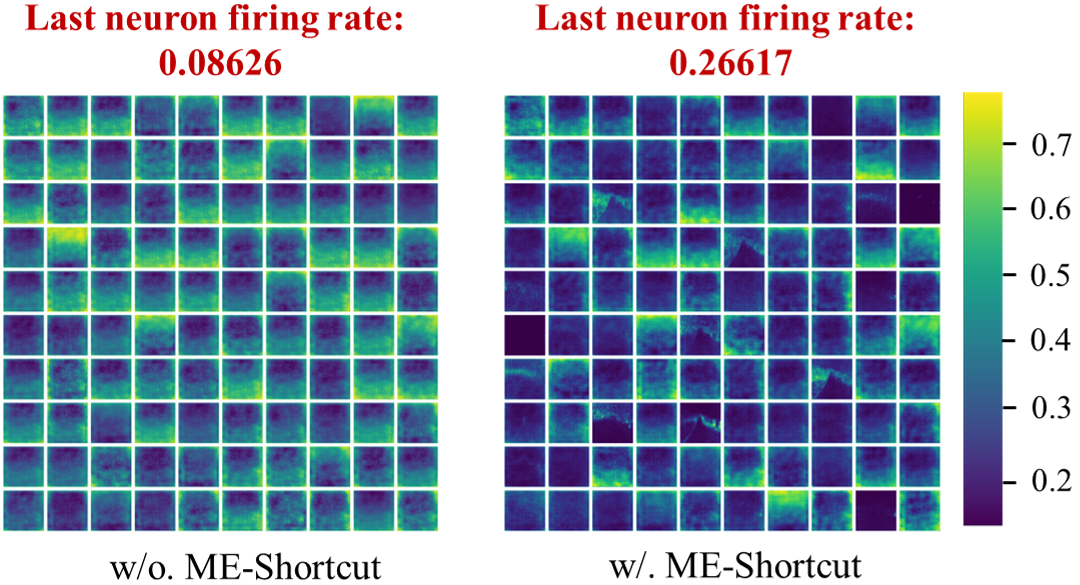} 
\caption{Segment Patterns Represents by Queries. We Average its corresponding segment predictions over the whole validation set of ADE20k. All the predictions are resized to the resolution of $200 \times 200$ for illustration purposes.}
\label{fig3}
\vspace{-0.2cm}
\end{figure}

\subsubsection{Information Deficiency in Query}
Preserving query information is crucial for effectively applying SNNs to Mask2Former. As demonstrated in Tab. \ref{tab:ablation}, removing the ME-Shortcut connection leads to a significant performance drop (approximately $-4.5\%$ mIoU). Fig. \ref{fig3} visualizes the average binary mask prediction for each query over the validation set of ADE20k, revealing that incorporating shortcut connections (ME-Shortcut) results in more sophisticated and distinct patterns within the mask embedding, indicating higher-quality binary mask predictions. Moreover, ME-Shortcut increases the firing rate of the final spiking neuron threefold (from 0.08626 to 0.26617). This suggests that the performance drop without ME-Shortcut is attributable to spike degradation, which our proposed shortcut effectively mitigates. Despite its conceptual simplicity, ME-Shortcut is a key design for the successful application of SNNs to complex architecture.

\subsubsection{NI-LIF Spiking Neuron}

Tab. \ref{tab:ablation} shows that directly applying I-LIF \cite{Xinhao} to Spike2Former results in sub-optimal performance (37.9\% mIoU compared to 46.3\%), primarily due to information deficiency in the cross-attention layer of transformer decoder. Normalizing activations within the cross-attention layer alone yields a substantial $+5.3\%$ mIoU improvement compared with I-LIF, highlighting the need for numerical stability and precise representation of Transformer-based model. Furthermore, when applying the NI-LIF to the whole network, the performance increases $8.4\%$. This improvement demonstrates the adaptability of NI-LIF in complex architecture and the effectiveness of NI-LIF in reducing the quantization error and enhancing the precise representation of features.

Further analysis in Tab. \ref{tab:segmentation} (Pascal VOC2012) reveals the impact of varying timesteps (T) and quantization steps (D). Increasing timesteps from T=1 to T=2 (with D=2) raises energy consumption from 50.6 mJ to 98.3 mJ with only $0.3\%$mIoU. Additionally, increasing quantization steps from D=2 to D=4 improves performance from 61.8\% to 75.1\% mIoU, with a moderate power increase from 50.6 mJ to 63.0 mJ. This suggests that increasing quantization steps can enhance the performance while maintaining reasonable energy consumption.

\section{Conclusion}
This work reduces the performance gap between Spiking Neural Networks (SNNs) and Artificial Neural Networks (ANNs) in image segmentation. The proposed Spike2Former introduces architectural innovations and spiking neuron optimizations, including two key modifications to address information deficiency. The Normalized Integer LIF (NI-LIF) mitigates information loss and enhances training stability by converting normalized integers into binary spikes. Spike2Former achieves state-of-the-art performance on three benchmark datasets, highlighting the potential of SNNs for complex segmentation tasks. Our analysis of spike degradation and information deficiency emphasizes the need to reduce information loss in SNNs for advanced architectures. This work lays a foundation for extending SNNs to dense prediction tasks with sophisticated designs.

\section{Acknowledgments}
This work was partially supported by CAS Project for Young Scientists in Basic Research (YSBR116), National Distinguished Young Scholars (62325603), National Natural Science Foundation of China (62236009, U22A20103, 62441606, 62406322), Beijing Science and Technology Plan
(Z241100004224011), Beijing Natural Science Foundation for Distinguished Young Scholars(JQ21015), China Postdoctoral Science Foundation (GZB20240824, 2024M753497), and Natural Science Foundation of China under Grant (62394314, 62394311).

\bigskip

\bibliography{reference}


\section*{Supplementary material}

# Spike2Former vs. Mask2Former

## Architecture design

Spike2Former adopts the same meta-architecture as Mask2Former (Cheng et al. 2022), which comprises a backbone for image feature extraction, a Feature Pyramid Network (FPN) pixel decoder with a Multi-scale Deformable self-attention transformer encoder for generating per-pixel embeddings, a stack of transformer decoders to process object queries in relation to the image features, and an embedding module to decode the per-pixel embeddings into binary mask predictions and class embeddings. Spike2Former introduces key modifications to the Multi-Scale Deformable Self-Attention transformer encoder and the mask embedding module, while making minor adjustments to the remaining components. The following section delves into the specific changes implemented within the transformer decoder and pixel decoder.

## Spike-Driven Transformer Decoder

**Transformer Decoder Design in Mask2Former** In Mask2Former, learnable object queries interact with multi-scale image features in a round robin fashion through a cross-attention mechanism. This strategy enables the queries to capture information about objects at various scales. After each Transformer decoder layer, Mask2Former employs an mask embedding module to generate binary mask predictions. These predictions are then converted into attention masks with a Sigmoid activation function and fed into the subsequent cross-attention layer, guiding the object queries to focus on regions containing segmented objects. However, this design proves less effective in the SNNs. The sparse nature of spike-based computation limits the ability of the attention mask, derived from the previous layer's binary predictions, to enhance the query's representation in the subsequent cross-attention layer. Furthermore, incorporating the mask attention can hinder the query's ability to effectively perceive and process the raw image features. Therefore, we omit this design, opting not to use mask attention within the cross-attention layers of our Spike2Former.

**Spike-Driven Self-Attention** The spike-driven self-attention has various variants. In this work, We take the inspiration of Meta-Spike2Former (Yao et al. 2024a) and proposed a spike-driven self-attention capable of dualing with sequence feature (Object Query). In Mata-Spikeformer, the calculation of self-attention can be written as:

$$\mathbf{Q_s} = SN(\text{RepConv}_1(\mathbf{U})), \quad (1)$$
$$\mathbf{K_s} = SN(\text{RepConv}_2(\mathbf{U})), \quad (2)$$
$$\mathbf{V_s} = SN(\text{RepConv}_3(\mathbf{U})), \quad (3)$$
$$SDSA(\mathbf{Q_s}, \mathbf{K_s}, \mathbf{V_s}) = SN(\mathbf{Q_s}(\mathbf{K_s^T V_s})), \quad (4)$$
$$\mathbf{U}' = \mathbf{U} + \text{RepConv}_4(SDSA(\mathbf{Q_s}, \mathbf{K_s}, \mathbf{V_s})). \quad (5)$$

where $U$ represents the input membrane potential, and $\text{RepConv}(\cdot)$ denotes the re-parameterization convolution operation with a kernel size of $3 \times 3$.

While this type of convolution has demonstrated significant energy efficiency gains with high-resolution images, we

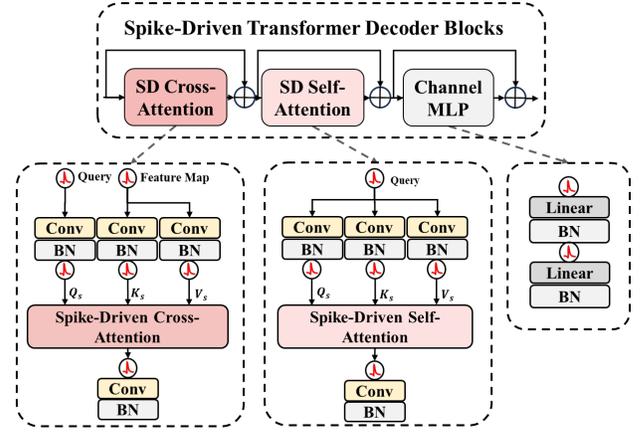

Figure 1: Illustration of the proposed Spike-Driven Transformer Decoder Blocks.

observed that it was not optimal for the specific demands of image segmentation. Therefore, to better suit the characteristics of this task, we replace the $3 \times 3$ re-parameterization convolution with a combination of a $1 \times 1$ standard convolution followed by Batch Normalization. This adjustment provides a more effective balance between computational efficiency and feature representation for accurate segmentation. The calculation of our Spike-Driven Self-Attention(SDSA) can be written as:

$$\mathbf{Q_s}, \mathbf{K_s}, \mathbf{V_s} = SN(\text{BN}(\text{Conv}(SN(\mathbf{U_{l-1}})))), \quad (6)$$
$$\mathbf{U}' = \text{BN}(\text{Conv}(SN(\mathbf{Q_s}\mathbf{K_s}^T\mathbf{V_s} * Scale))) \quad (7)$$

where the $scale$ of SDSA can be re-parameterized into the spiking neuron's threshold.

**Spike-Driven Cross-Attention** The calculation of spike-driven cross-attention are similar to the self-attention operation. Moreover, the $K_s$ and $V_s$ in cross-attention are obtained from the multi-scale image features generated from the pixel decoder. We calculate the spike-driven cross-attention as follow:

$$\mathbf{Q_s} = SN(\text{BN}(\text{Conv}(SN(\mathbf{U_{l-1}})))), \quad (8)$$
$$\mathbf{K_s}, \mathbf{V_s} = SN(\text{BN}(\text{Conv}(SN(\mathbf{F_i})))), i = 1, 2, 3, \quad (9)$$
$$\mathbf{U}' = \text{BN}(\text{Conv}(SN(\mathbf{Q_s}\mathbf{K_s^T}\mathbf{V_s} * Scale))) \quad (10)$$

Where $\mathbf{F_1}, \mathbf{F_2}, \mathbf{F_3}$ indicates the multi-scale feature maps in $\{\frac{1}{16}, \frac{1}{8}, \frac{1}{4}\}$, The $scale$ of SDCA can be re-parameterized into the spiking neuron's threshold.

## Feature Pyramid Network Pixel Decoder

In Spike2Former, we introduce a subtle yet efficient modification to the FPN pixel decoder. Specifically, we substitute the $3 \times 3$ channel convolution in the original Spike-FPN with a more computationally efficient $3 \times 3$ depth-wise convolution. This adjustment aims to reduce the energy consumption associated with generating per-pixel embeddings.

As evidenced by the results presented in Table 1, this alteration yields a notable improvement in energy efficiency. We observe a significant reduction in energy consumption

| Method | Power(mJ) | mIoU(%) |
|---|---|---|
| Vanilla SpikeFPN | 95.8 | 45.9 |
| DWConv1x1 SpikeFPN | 68.0 | 46.3 |

Table 1: Ablation study on the depth-wise convolution in pixel decoder in ADE20k dataset. We equipped Spike2Former with different pixel decoder, reporting its energy consumption and performance.

($-27.8$ mJ) while simultaneously achieving a slight performance gain of approximately $0.4\%$ mIoU. This outcome highlight the effectiveness of this modification in balancing computational cost and segmentation accuracy.

## Dataset and Training Settings

In this section, we presents the detail of dataset, the training setting on each dataset, and the theoretical energy consumption calculation method.

### Semantic segmentation

We conduct semantic segmentation on three popular datasets: ADE20k(Zhou et al. 2017), CityScapes(Cordts et al. 2016), and Pascal VOC2012(Everingham et al. 2010). The evaluation matrix are mean Interact of Union(mIoU).

**ADE20k** ADE20k contains 20,000 images for training and 2,000 images for validation. The dataset is a subset of ADE20k-Full, featuring 150 semantic categories. In this work, we validate the Spike2Former with an input resolution of $512 \times 512$ and a learning rate of 0.0002. We employ the data augmentation strategy as described in (Cheng et al. 2022). During training, the images are resized so that the shortest side does not exceed 512 pixels. We train the model on ADE20k for 160k iterations with a batch size of 36.

**CityScapes** CityScapes is an urban, egocentric street view dataset with high-resolution images ($1024 \times 2048$ pixels) across 19 semantic categories. We use 2,975 images for training, 500 images for validation, and 1,525 images for testing. During training, we crop the images to $512 \times 1024$ and apply data augmentation using the same strategy as in (Cheng et al. 2022). We train the model for 90k iterations with a batch size of 32. During inference, we resize the longer side of the images to 2048 pixels and use multi-scale testing as (Cheng, Schwing, and Kirillov 2021).

**Pascal VOC 2012** The Pascal VOC 2012 dataset comprises approximately 11,530 images, of which 5,717 are used for training and 5,823 for testing. The image resolutions vary, typically ranging from 300x500 to 500x400 pixels. The dataset includes 20 semantic categories such as people, animals, vehicles, and furniture. The scenes mainly depict everyday natural environments, including streets, homes, and outdoor settings. We use an input resolution of $512 \times 512$ with a batch size of 32 for training and resize the longer side of the images to 512 pixels during inference.

### Panoptic segmentation

Panoptic segmentation is a more challenging segmentation task that provides comprehensive scene understanding by simultaneously performing semantic and instance segmentation. The core requirement is to assign every pixel a semantic label, such as "road" or "tree," while also distinguishing between different instances of objects within the same category, like multiple "cars." This task is particularly challenging because it must effectively combine instance-level and pixel-level segmentation into a single, coherent output, a contrast to semantic segmentation, which does not differentiate between individual instances of the same class. Panoptic segmentation is essential for advanced applications like autonomous driving, where precise scene interpretation is critical. *To the best of our knowledge, no panoptic segmentation results have been reported in SNNs. In this work, we present the first panoptic segmentation results in SNNs on the widely-used COCO2017 panoptic dataset.* We evaluate the performance with Panopic Quality(PQ) in 'All' object, 'Thing', and 'Stuff', denoting as $PQ_{All}$, $PQ_{Thing}$, and $PQ_{Stuff}$, separately.

**COCO Panoptic** COCO panoptic is one of the most widely used datasets for panoptic segmentation, containing 133 categories (80 "thing" categories with instance-level annotations and 53 "stuff" categories). It includes 118,000 images for training and 5,000 images for validation. All images are sourced from the COCO2017 dataset. We employ large-scale jittering augmentation as described in (Du et al. 2021) and crop the images to $1024 \times 1024$ during training.

### Theoretical Energy Consumption

The power consumption of ANNs and SNNs can be estimated using the following equations:

$$E_{ANN} = O^2 \times C_{in} \times C_{out} \times k^w \times E_{MAC}, \quad (11)$$

$$E_{SNN} = (T \times D) \times R_{LIF} \times O^2 \quad (12)$$

$$\times C_{in} \times C_{out} \times k^2 \times E_{AC}. \quad (13)$$

Here, $O$ represents the output size, $C_{in}$ and $C_{out}$ denote the number of input and output channels respectively, $k$ is the kernel size, $R_{LIF}$ represents the firing rate of the LIF spiking neuron, $T$ is the timestep, and $D$ is the upper limit of integer-activation during training.

We adopt a widely used energy consumption evaluation method for SNNs(Yao et al. 2024b,a; Wang et al. 2023), assuming a 32-bit floating-point implementation on a 45nm technology. In this context, $E_{MAC} = 4.6pJ$ and $E_{AC} = 0.9pJ$ represent the energy consumed per MAC (Multiply-ACcumulate operations) and AC (ACcumulate operations) operation respectively (Yao et al. 2024b).

As shown in Eq. 13, the reduced power consumption of SNNs stems from their reliance on sparse addition operations, reflected in the use of $E_{AC}$.

### Training Setting

We conducted our experiments using the popular codebases MMSegmentation (Contributors 2020) and MMDetection (Chen et al. 2019), which are widely used for semantic

| Method | Model | PQ All(%) | Thing(%) | Stuff(%) | Power (mJ) | Params (M) |
|---|---|---|---|---|---|---|
| ANN | MaskFormer | 46.5 | 51.0 | 39.8 | 832.6 | 45 |
| | | 47.7 | 51.7 | 41.7 | 823.4 | 42 |
| | Mask2Former | 51.9 | 57.7 | 43.0 | 1039.6 | 44 |
| | | 53.2 | 59.3 | 44.0 | 1067.2 | 47 |
| SNN | SpikeFPN* | 14.5 | 9.8 | 21.9 | 448.0 | 36 |
| | Spike2Former(**Ours**) | **30.9** | **31.7** | **30.0** | **222.6** | 34 |

Table 2: The performance of panptic segmentation on COCO2017 with 133 categories. We use $T \times D = 1 \times 4$ in COCO2017 panoptic. The result of MaskFormer and Mask2Former are equipped with R50(above) and Swin-t(Below) as backbone. * indicates the self-implement result (Yao et al. 2024a).

| Model | mIoU(%) | $T \times D$ | Power(mJ) |
|---|---|---|---|
| Spike2Former | 5.2* | 1×1 | 60.2 |
| | 61.8 | 1×2 | 50.6 |
| | 62.1 | 2×2 | 98.3 |
| | 75.1 | 1×4 | 63.0 |
| | 75.3 | 2×4 | 117.4 |
| | 75.4 | 4×4 | 221.9 |
| | 76.0 | 1×8 | 66.9 |

Table 3: Ablation study on the various $T \times D$ in Pascal VOC2012 with input resolution of $512 \times 512$. * indicates the setting is not work well because of the huge information deficiency.

and panoptic segmentation in the context of ANNs, respectively. To enable the use of these versatile frameworks within the SNN domain, we introduced modifications to support SNN training. These adaptations pave the way for efficient development of future SNN-based segmentation models. All the experiment are train and evaluate on 4 NVIDIA A100 GPUs.

## Experiment Result on Panoptic Segmentation

Tab. 2 presents a comprehensive comparison of Spike2Former against state-of-the-art ANN and SNN methods on the COCO2017 panoptic segmentation benchmark, considering metrics such as mIoU, parameter count, and power consumption. For this evaluation, we re-implemented Spike-FPN (Yao et al. 2024a) within our Spike-mmdetection codebase.

Our proposed Spike2Former significantly advances the performance limits of SNNs for this task. We achieve $30.9\%$ PQ, $31.7\%$ PQ, and $75.4\%$ PQ for the "All," "Things," and "Stuff" categories, respectively. These results represent substantial improvements of $+16.4\%$, $+21.9\%$, and $+8.1\%$ over the previous SNN state-of-the-art, Spike-FPN (Yao et al. 2024a). Furthermore, Spike2Former exhibits considerable advantages in energy efficiency compared to existing SNNs. Specifically, Spike2Former attains a $PQ_{All}$ of $30.9\%$ with a power consumption of **222.6mJ**, while SpikeFPN (Yao et al. 2024a) achieves $33.6\%$ $PQ_{All}$ at the cost of $448.0 mJ$.

While there remains a performance gap between Spike2Former and ANN methods, our approach demonstrates a more energy-efficient way to perform panoptic segmentation. For instance, Spike2Former exhibits a $4.8\times$ improvement in energy efficiency compared to Mask2Former with Swin-t as backbone.

## Ablation Study
### Ablation Study on T and D

We conducted an ablation study to investigate the effects of varying timesteps ($T$) and quantization steps ($D$). As illustrated in Table 3, increasing $T$ from 1 to 2 (while keeping $D$ constant at 2) leads to a considerable rise in energy consumption, from 50.6 mJ to 98.3 mJ. Conversely, increasing $D$ from 2 to 4 (with $T$ fixed at 1) yields a substantial performance boost, raising the mIoU from $61.8\%$ to $75.1\%$, while incurring only a moderate power increase from 50.6 mJ to 63.0 mJ. This suggests that increasing the quantization steps is an effective strategy for enhancing performance while maintaining reasonable energy efficiency.

Notably, we observed a significant performance drop when the virtual timestep was set to 1. We hypothesize that this is because a pure 0/1 spike sequence is not well-suited for the Mask2Former architecture, adversely affecting numerical stability and hindering effective interactions between features. Therefore, we recommend using a virtual timestep of $2 \sim 4$ to normalize the output into integer spikes for complex models.

### Visualization

As shown in Fig. 2, Fig. 3, and Fig. 4, we visualize the semantic segmentation results of Spike2Former on the ADE20k, CityScapes, and Pascal VOC2012 datasets. Our proposed Spike2Former demonstrates remarkable capabilities in identifying small objects, accurately classifying object categories, and producing precise segmentation along object boundaries.

Figure 2: Visualization of semantic segmentation predictions on the ADE20K dataset: Spike2Former with Meta-Spikeformer backbone which achieves **46.3** mIoU on the validation set. First and third columns: ground truth. Second and fourth columns: prediction. Last row shows failure cases

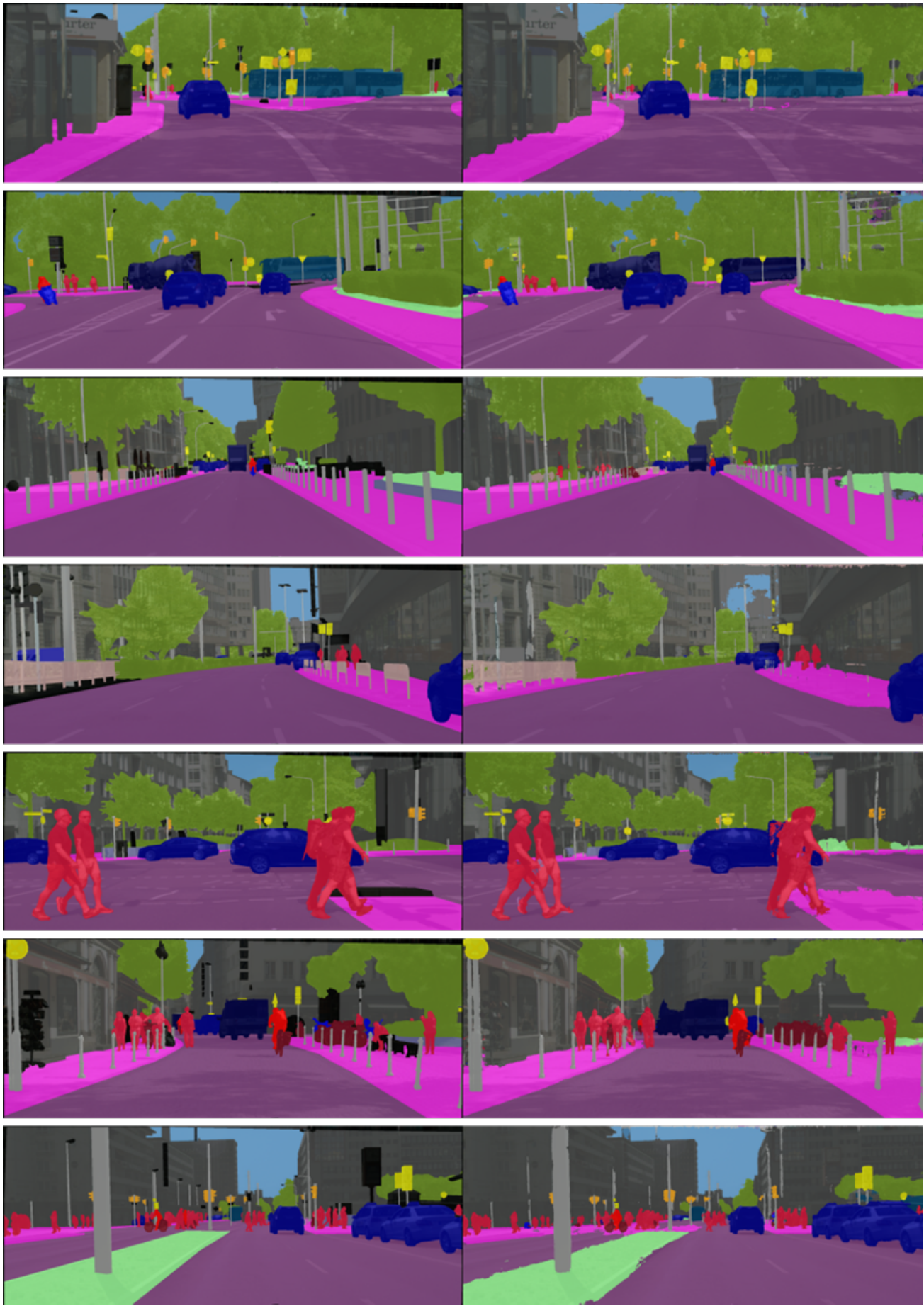

Figure 3: Visualization of semantic segmentation predictions on the CityScapes dataset: Spike2Former with Meta-Spikeformer backbone which achieves **75.2** mIoU on the validation set. First columns: ground truth. Second columns: prediction. Last row shows failure cases

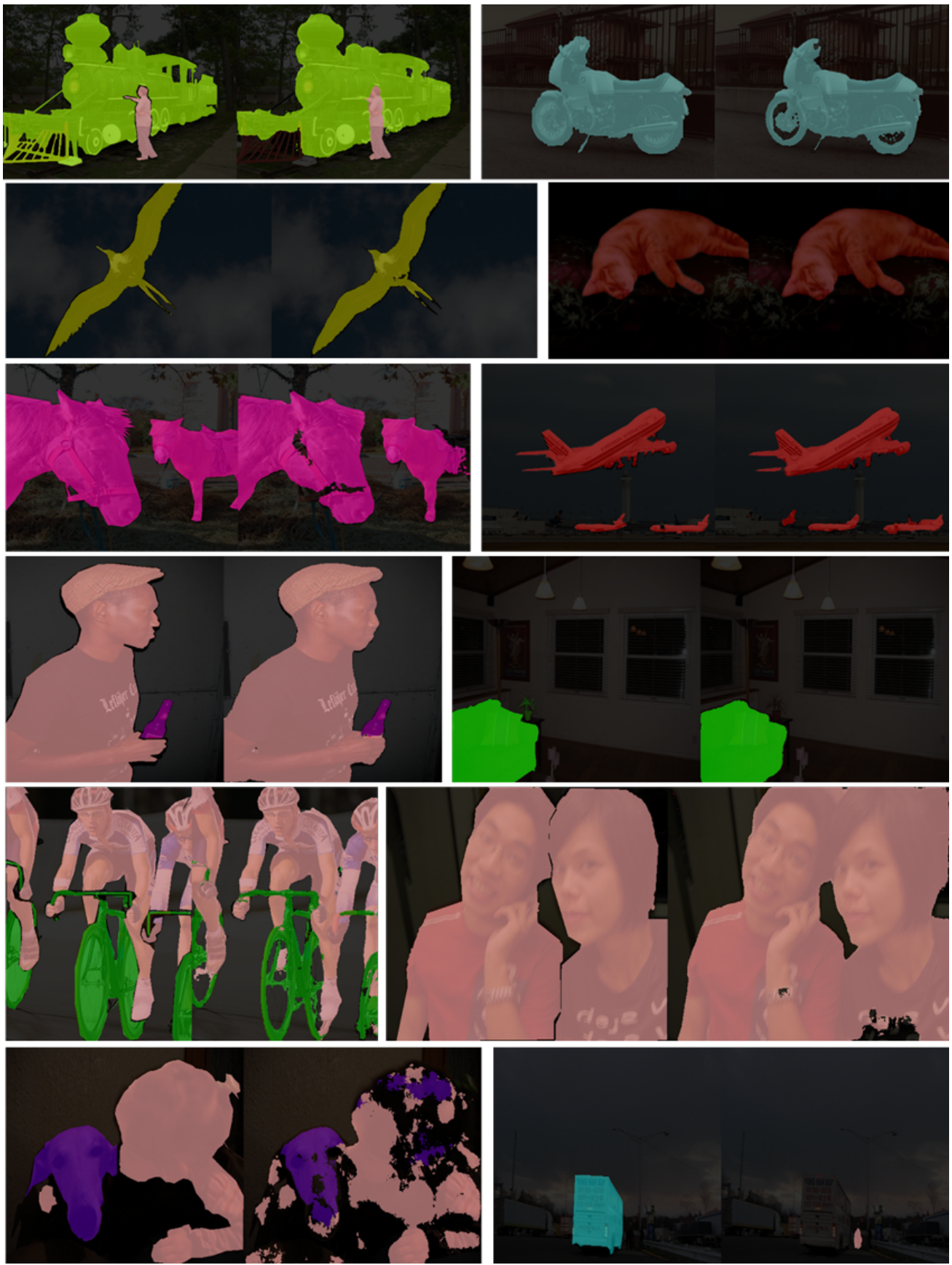

Figure 4: Visualization of semantic segmentation predictions on the Pascal VOC2012 dataset: Spike2Former with Meta-Spikeformer backbone which achieves **75.4** mIoU on the validation set. First and third columns: ground truth. Second and fourth columns: prediction. Last row shows failure cases



\end{document}